\documentclass[letterpaper, 10 pt, conference]{ieeeconf}
\usepackage{cite}
\usepackage{amsmath,amssymb,amsfonts} 
\usepackage{graphicx}
\usepackage{textcomp} 
\usepackage{multirow}
\usepackage{gensymb}
\usepackage{xcolor} 
\usepackage[ruled,vlined]{algorithm2e}

\IEEEoverridecommandlockouts                              

\newcommand{\veryshortarrow}[1][3pt]{\mathrel{%
   \vcenter{\hbox{\rule[-.2pt]{#1}{.4pt}}}%
   \mkern-4mu\hbox{\usefont{U}{lasy}{m}{n}\symbol{41}}}}

\makeatletter

\setbox0\hbox{$\xdef\scriptratio{\strip@pt\dimexpr
    \numexpr(\sf@size*65536)/\f@size sp}$}

\newcommand{\scriptveryshortarrow}[1][3pt]{{%
    \vcenter{\hbox{\rule[\scriptratio\dimexpr-.2pt\relax]
               {\scriptratio\dimexpr#1\relax}{\scriptratio\dimexpr.4pt\relax}}}%
   \mkern-4mu\hbox{\let\f@size\sf@size\usefont{U}{lasy}{m}{n}\symbol{41}}}}

\newcommand{\etal}{et al.}

\makeatother

\newcommand{\figfront}{
 \begin{figure}[htbp]
\includegraphics[width=\linewidth]{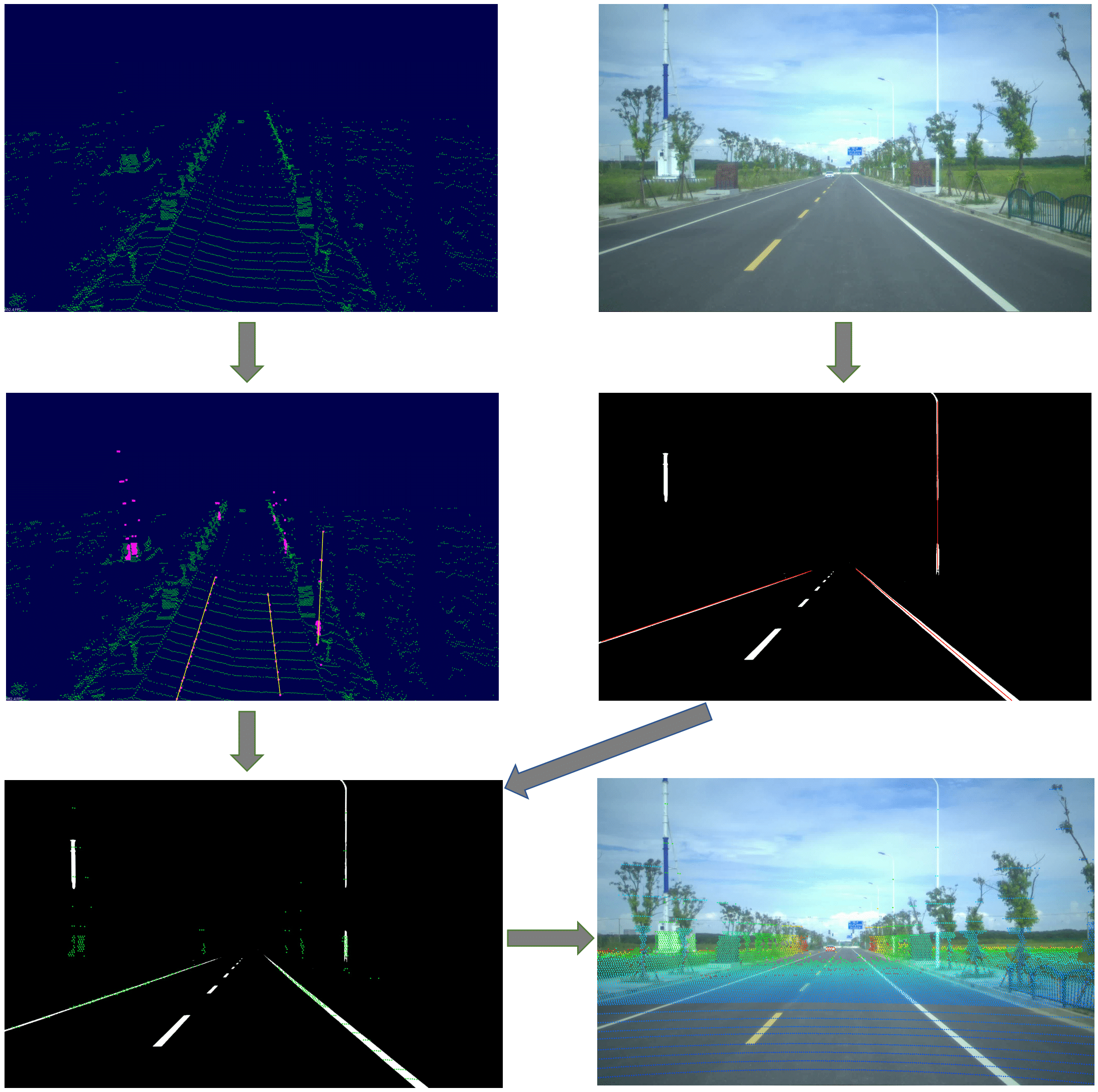}
\caption{The proposed CRLF extracts straight line features from both the image (white region) and point cloud (pink points), and estimates the coarse calibration by solving the P3L problem under line feature constraints. The final extrinsic parameters after refinement with semantic line features is used to project all point cloud data to the original image.}
\vspace{-3mm}
\label{fig:front_pic}
\end{figure}
}

\newcommand{\figqualitativee}{
\begin{figure*}[htbp]
\centering
\includegraphics[width=1.\textwidth]{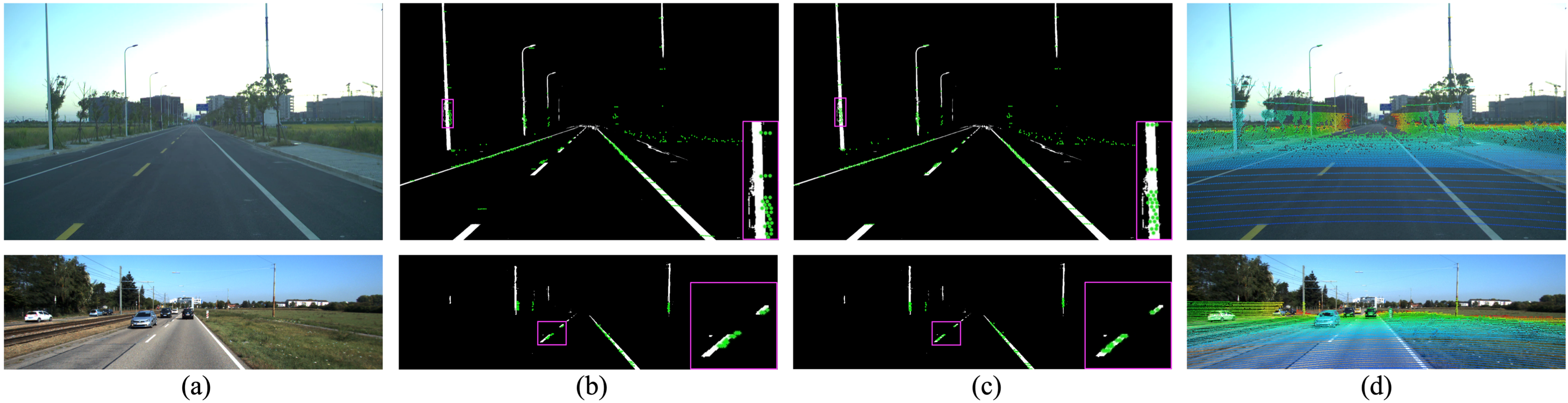}
\vspace{-6mm}
\caption{Qualitative results of CRLF. The 1st row are the results on our in-house dataset, and the 2nd row are the results on KITTI dataset. (a): Original images from the camera. (b): Projection results of the line features by the coarse calibration. (c): Projection results of the line features by the refined calibration. (d): Projection results of the whole point cloud  on the original image by the refined calibration. The alignment of local regions are marked with pink box and magnified on the right bottom to better visualize the quality of the calibration in (b) and (c).} 
\vspace{-2mm}
\label{fig:qualitative}
\end{figure*}
}

\newcommand{\figcrf}{
 \begin{figure}[htbp]
\includegraphics[width=1.\linewidth]{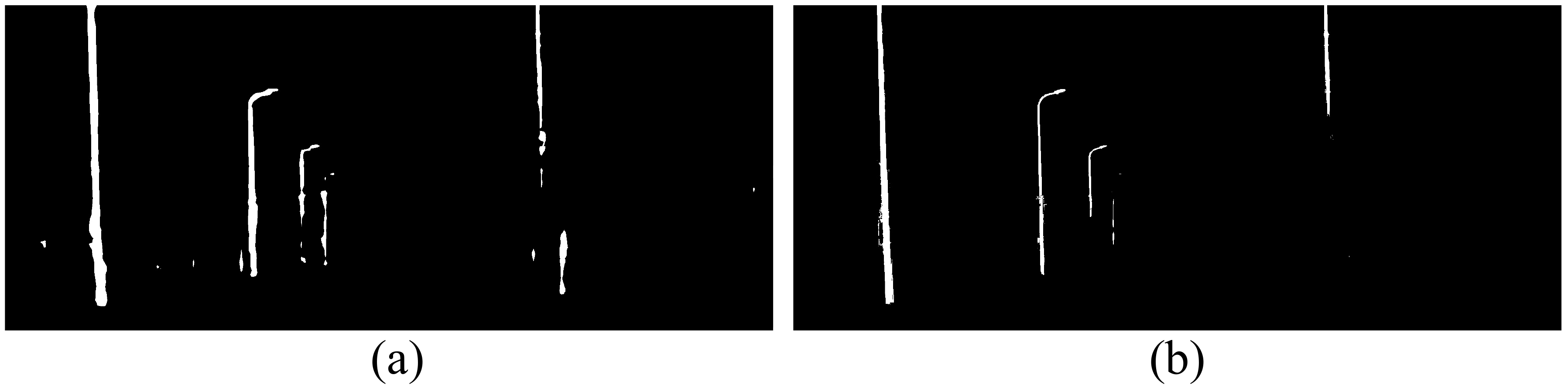}
\vspace{-6mm}
\caption{Pole masks before CRF operator (a)  and after CRF operator (b). }
\vspace{-4mm}
\label{fig:crf}
\end{figure}
}

\newcommand{\figcoarse}{
 \begin{figure}[tbp]
\includegraphics[width=\linewidth]{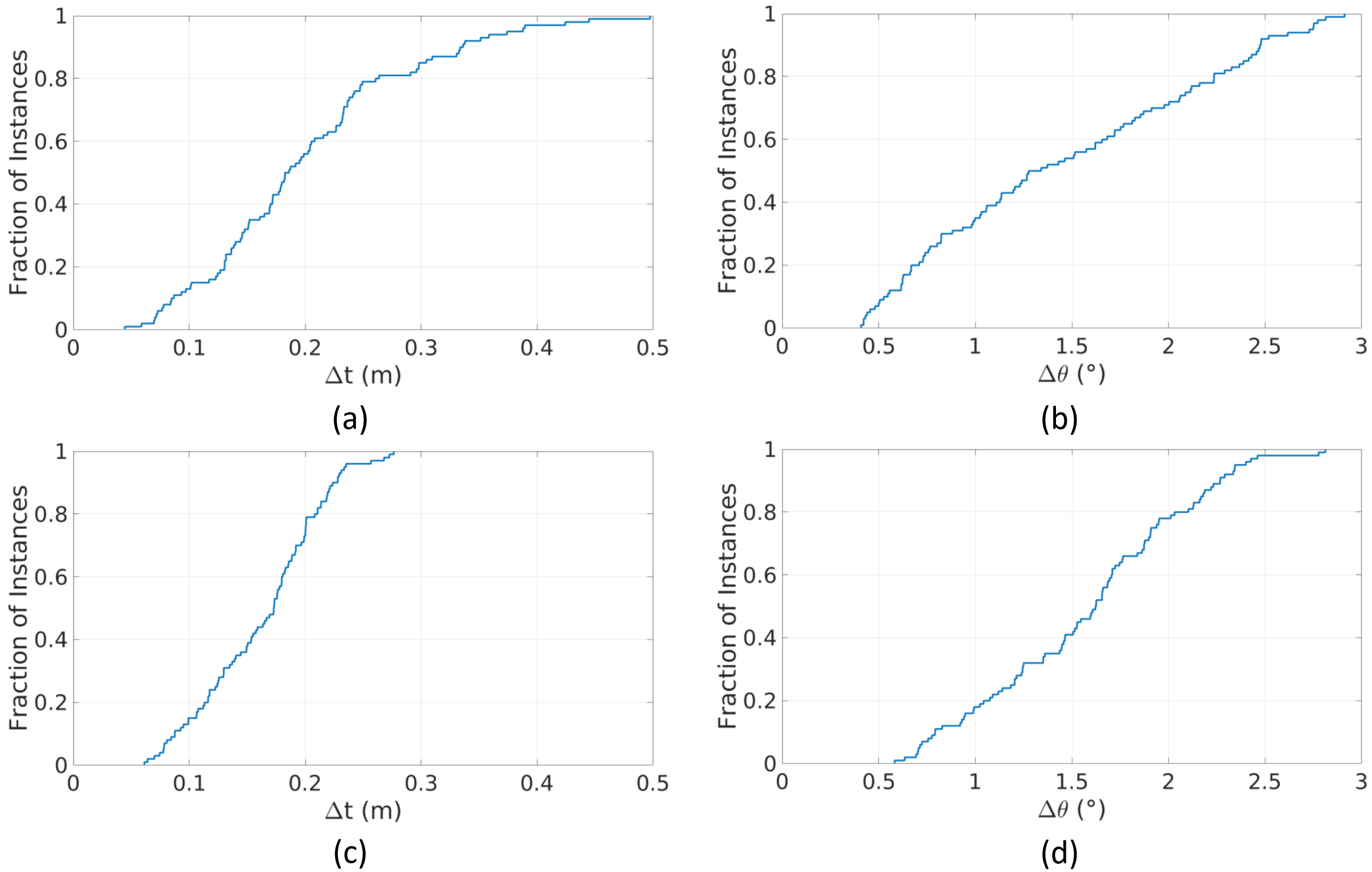}
\vspace{-4mm}
\caption{Results of the coarse calibration. (a),(b) show the cumulative distribution of the translation and rotation errors on KITTI dataset. (c),(d) show those on our in-house dataset.}
\label{fig:coarse}
\end{figure}
}

\newcommand{\figrefine}{
 \begin{figure}[tbp]
\includegraphics[width=\linewidth]{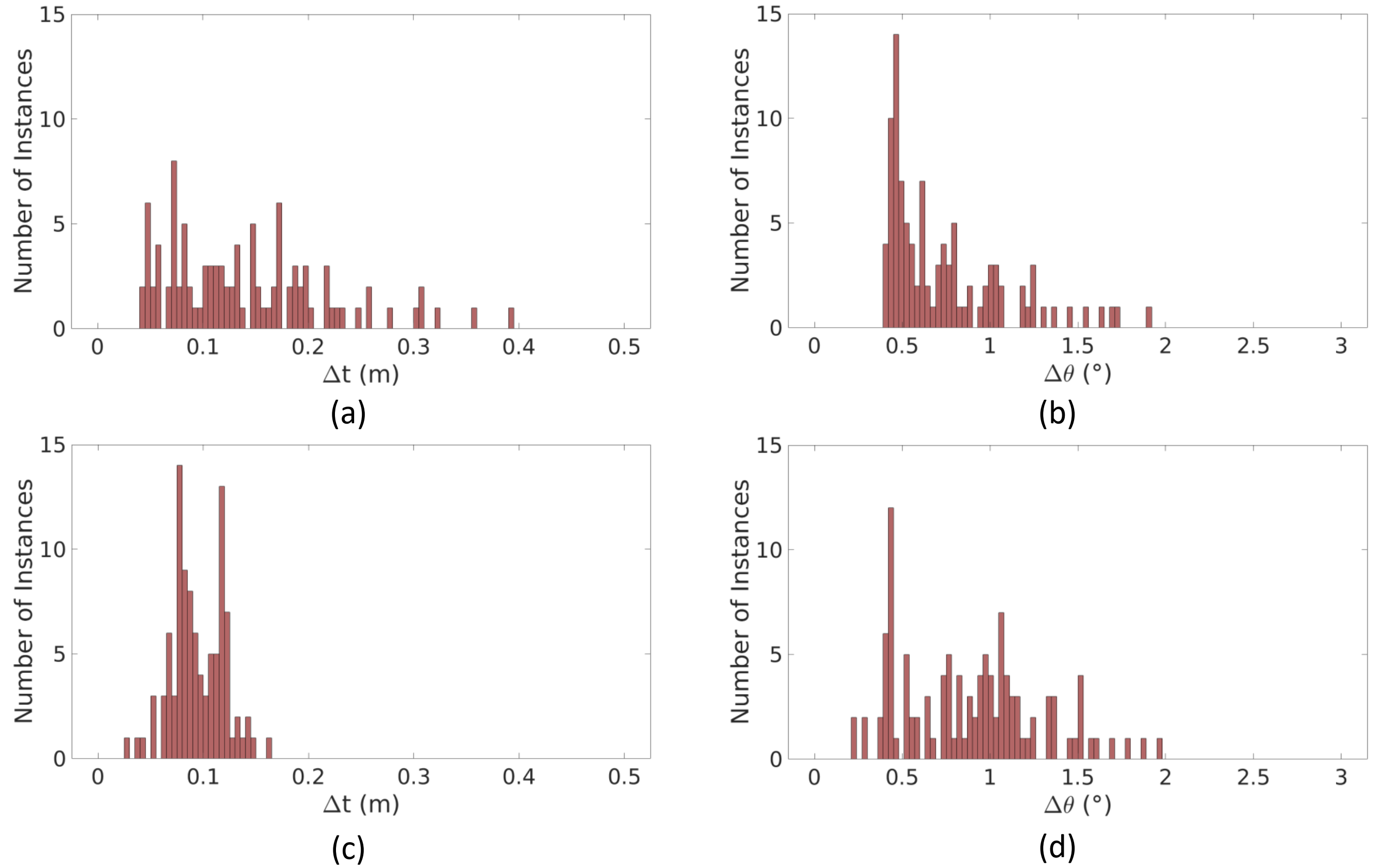}
\vspace{-6mm}
\caption{Results of the refined calibration. (a),(b) show the histogram of the translation and rotation errors on KITTI dataset. (c),(d) show those on our in-house dataset.}
\label{fig:refine}
\end{figure}
}

\newcommand{\figconverge}{
 \begin{figure}[tbp]
\includegraphics[width=\linewidth]{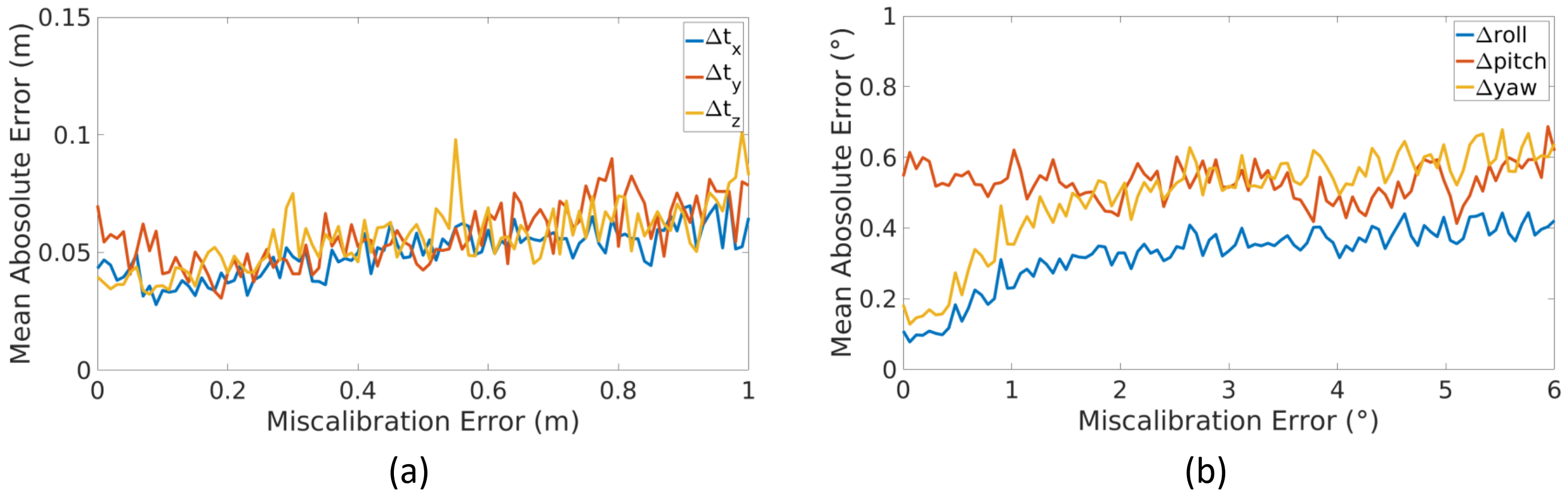}
\vspace{-6mm}
\caption{MAE of the refined calibration versus the initial miscalibration error. (a) shows the MAE of translation, and (b) shows the MAE of rotation. }
\label{fig:converge}
\end{figure}
}

\newcommand{\figpnl}{
\begin{figure}[htbp]
\centering
\includegraphics[width=0.9\linewidth]{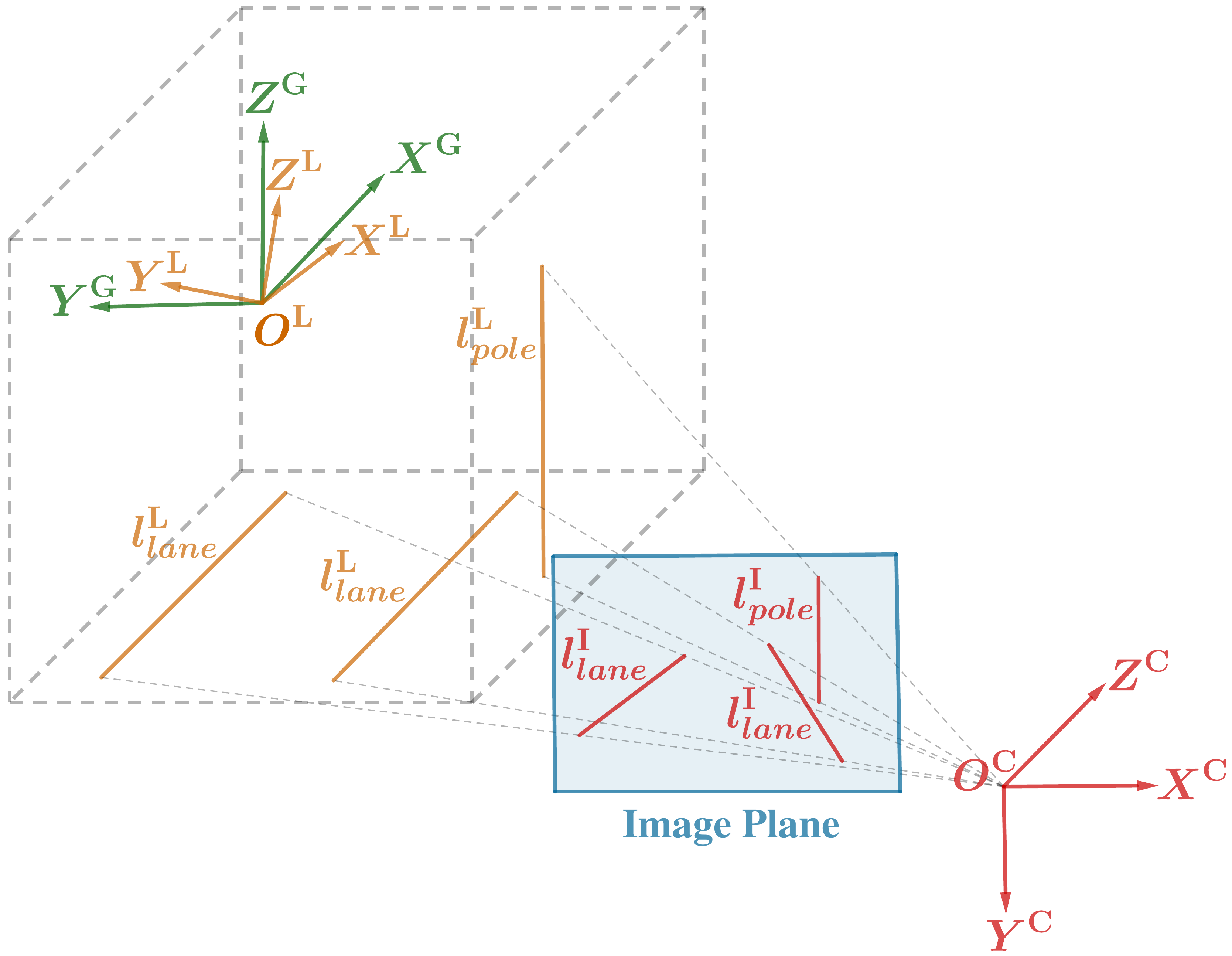}
\vspace{-3mm}
\caption{Illustration of the coarse calibration. The yellow and red lines are fitted from the line features in the point cloud and the image respectively. The coarse calibration is given by solving the P3L problem to project orange lines to the corresponding red lines in image plane, with the help of  an intermediate  ground-parallel coordinate system $\mathrm{G}$ marked in green. }
\vspace{-5mm}
\label{fig:p3l}
\end{figure}
}
\newcommand{\tableEfficiency}{
\begin{table}[htbp]
\caption{Average Running Time for Different Steps of CRLF }
\vspace{-3mm}
\centering
\begin{tabular}{l|l}
Lane Feature Extractor &  0.104s \\ 
Coarse Calibration     &  0.032s\\ 
Calibration Refinement &  0.175s
\end{tabular}  
\label{table:efficiency} 
\end{table}
}

\newcommand{\tableResult}{
\begin{table}[tbp]
\centering
\caption{MAE for Translation and Rotation of CRLF} 
\vspace{-3mm}
 \setlength{\tabcolsep}{2pt}
\begin{tabular}{|c|c|c|c|c|c|c|c|}
\hline
Dataset                & Calib   & $\Delta t_x$(m) & $\Delta t_y$(m) & $\Delta t_z$(m) & $\Delta$\textit{roll}(\degree) &  $\Delta$\textit{pitch}(\degree) &  $\Delta$\textit{yaw}(\degree) \\ \hline
\multirow{2}{*}{KITTI} & Coarse  & 0.121             &  0.067            &    0.182          &  0.628     & 1.043      & 0.805     \\ \cline{2-8} 
                       & Refined & 0.082             &          0.046    & 0.097             &  0.216     & 0.546       & 0.492     \\ \hline
\multirow{2}{*}{Ours}  & Coarse  &   0.053           &  0.117            &      0.074        & 0.943     & 1.092      & 0.684    \\ \cline{2-8} 
                       & Refined &  0.018            &     0.069         &        0.015      &   0.332   &    0.613    & 0.395    \\ \hline
\end{tabular}  

\label{table:result}
\end{table}
}

\newcommand{\alg}{
\setlength{\textfloatsep}{5pt}
\begin{algorithm}[htbp]
\SetAlgoLined
\KwIn{initial rotation $\mathbf{r_0}$, initial translation $\mathbf{t_0}$, initial step size $\eta$, step size lower bound $\eta_0$, max search count $\mathrm{\max\_cnt}$, decay factor $\mathrm{k}$}
\KwOut{refined rotation $\hat{\mathbf{r}}$, refined translation $\hat{\mathbf{t}} $} 
 
 $\hat{\mathbf{r}} = \mathbf{r_0},\hat{\mathbf{t}} = \mathbf{t_0}$\;
 $\mathrm{J_{max}} =  \boldsymbol{\mathcal{J}}(\mathbf{R}(\mathbf{r_0}),\mathbf{t_0})$\;
 \While{$\eta \geq \eta_0$}{
 $\mathrm{cnt} = 0$\;
  \While{$\mathrm{cnt} < \mathrm{max\_cnt}$}{
 Sample $\delta\theta$ from $[-\theta_0, \theta_0]$\;
 Sample $\delta t$ from $[-t_0, t_0]$\;
 Sample vector $\mathbf{v}$ from the unit sphere\; 
 $\delta\mathbf{r} = \delta\theta \cdot \mathbf{v}$\;  
 $\mathrm{J_{current}} =  \boldsymbol{\mathcal{J}}(\mathbf{R}(\eta\cdot \delta \mathbf{r})\cdot \mathbf{R}(\hat{\mathbf{r}}),\eta\cdot\delta \mathbf{t} + \hat{\mathbf{t}})$\;
  \If{$\mathrm{J_{current}}>\mathrm{J_{max}}$}{
    $\hat{\mathbf{r}} = \mathbf{R}^{-1}(\mathbf{R}(\eta \cdot\delta \mathbf{r})\cdot \mathbf{R}(\hat{\mathbf{r}}))$; $\hat{\mathbf{t}} = \eta\cdot\delta \mathbf{t} + \hat{\mathbf{t}}$\;
    $\mathrm{J_{max}} =  \mathrm{J_{current}}$\; 
   } 
   $\mathrm{cnt}$++\;
   } 
    $ \eta = \mathrm{k} \cdot \eta$ \; 
 }
 \caption{Extrinsic Calibration Refinement}
 \label{alg:1}
 
\end{algorithm}
 
}

\title{\LARGE \bf
CRLF: Automatic Calibration and Refinement based on Line Feature for LiDAR and Camera in Road Scenes
}

\author{Tao Ma$^*$, Zhizheng Liu$^*$, Guohang Yan, and Yikang Li$^{\dagger}$ 
\thanks{$^{*}$ Equally contributed to the work.}
\thanks{$^{\dagger}$ Corresponding author.}
\thanks{Tao Ma, Zhizheng Liu, Guohang Yan, and Yikang Li are with Autonomous Driving Group, SenseTime, Shanghai, China. {\tt\small \{matao, liuzhizheng, yanguohang, liyikang\}@senseauto.com}}
}
    
\begin{document}
 
\maketitle

\begin{abstract}
For autonomous vehicles, an accurate calibration for LiDAR and camera is a prerequisite for multi-sensor perception systems.
However, existing calibration techniques require either a complicated setting with various calibration targets, or an initial calibration provided  beforehand, which greatly impedes their applicability in large-scale autonomous vehicle deployment. To tackle these issues, we propose a novel  method to  calibrate the extrinsic parameter for LiDAR and camera in road scenes. Our method introduces line features from static straight-line-shaped objects such as road lanes and poles in both image and point cloud and formulates the initial calibration of extrinsic parameters as a perspective-3-lines~(P3L) problem.
Subsequently, a cost function defined under the semantic constraints of the line features is designed to perform refinement on the solved coarse calibration. The whole procedure is fully automatic and user-friendly without the need to adjust environment settings or provide an initial calibration. We conduct extensive experiments on KITTI and our in-house dataset, quantitative and qualitative results demonstrate the robustness and accuracy of our method.
 \end{abstract}

\section{INTRODUCTION}
Autonomous driving has recently become one of the most popular technologies in both industry and academia. As a complicated system, it requires numerous modules collaboratively to work together. Among them, the perception system is the most challenging and upstream part. 
Since different sensors have their advantages and shortcomings, fusing multiple heterogeneous sensors becomes the key to robust and accurate perception ability. 
As the most widely used sensors, LiDAR and camera are usually equipped to achieve a dense 3D perception. 
Therefore, to fuse their strengths for better perception performance, an accurate extrinsic calibration for LiDAR and camera serves as the cornerstone of this multi-sensor perception system.

Researchers have investigated various scenarios to improve the accuracy and efficiency of calibration results, such as specific targets like checkerboards\cite{geiger2012automatic,wang2017reflectance,zhou2018automatic,an2020geometric} or ordinary boxes\cite{pusztai2017accurate}, arbitrary edge feature\cite{levinson2013automatic}, and semantic objects like cars\cite{zhu2020online}. Among these methods, automatic target-less calibration methods\cite{levinson2013automatic, zhu2020online} are user-friendly ones, since they can cut down time consumption and manual work. Besides, they greatly increase the flexibility of the calibration settings and the possibility for online calibration.

\figfront


However, these methods usually need a provided initial calibration due to the lack of effective constraint information, which adds additional effort and impedes its practice in large-scale applications. 
To mitigate this issue, \cite{wang2020soic} propose a fully automatic approach  based on semantic constraints. Specifically, they use semantic centroids~(SCs)  for each semantic class, i.e., pedestrians, cyclists, and especially vehicles to conduct the initial calibration.
However, multiple-frame inputs are compulsory to extract enough SCs, and these SCs usually lie near the ground plane, 
which leads to considerable difficulties in solving the perspective-n-points~(PnP) problem to give a robust initial calibration.
Additionally, these types of objects are usually dynamic in road scenes, which would produce uncorrectable motion distortion of laser points.
Therefore, to enhance the accuracy and stability of the fully automatic methods, a vital step is to extend the on-road objects to  static ones with sufficient spatial constraints in road scenes.

To this end, we leverage straight-line-shaped objects including lanes and poles to enrich our potential targets for calibration. Correspondingly, this new calibration method is named CRLF: automatic Calibration and Refinement based on Line Feature. Firstly, straight-line features are extracted from road lanes and poles of a single pair of image and point cloud. Then we formulate the calibration of extrinsic parameters as a perspective-3-lines~(P3L) problem. As shown in Fig.~\ref{fig:front_pic}, CRLF could segment line-shaped targets in both the image and the point cloud, in which three straight-line correspondences can provide enough constraints to compute a coarse calibration with appreciable speed.
Subsequently, we supervise the matching between those image masks and segments of the point cloud through a cost function to further refine the calibration result.
Considering the fewer requirements for environment settings and the fully-automatic calibration procedure, the proposed method could contribute to a more efficient and practical  large-scale autonomous vehicle~(AV) and deployment and test.


The contributions of this work are summarized as four-fold: 
\begin{enumerate}
\item We propose a fully automatic and target-less method, CRLF, for LiDAR-Camera extrinsic calibration based on straight-line-shaped objects in road scenes.
\item A \textit{line feature extractor} is proposed to extract straight line features of lanes and poles for both the image and the point cloud.
\item We introduce a more robust initial calibration estimation algorithm based on line correspondences and a newly defined cost function for calibration refinement.
\item Evaluated on KITTI and our in-house dataset, we demonstrate CRLF's robustness and accuracy in quantitative and qualitative results.
\end{enumerate}

\section{RELATED WORK}
Early methods often rely on calibration targets that can be observed from both the image and the point cloud. The most common ones are checkerboards, and \cite{geiger2012automatic,wang2017reflectance,zhou2018automatic,an2020geometric} used points, lines, planes, and intensity correspondences on checkerboards. Researchers also design targets that are tailored to the task of extrinsic calibration to better capture the feature correspondences\cite{pusztai2017accurate,park2014calibration,velas2014calibration,cai2020novel}. However, these methods usually need a complicated setup for placing these  targets and are laborious to users or autonomous driving companies with lots of demands on calibration.

To overcome this problem, researchers have also proposed methods that don't need calibration targets and can work in arbitrary scenes. Pandey \etal \cite{pandey2012automatic} suggested using mutual information to match grayscale intensity in image and reflectance intensity in the point cloud, and which is further improved by considering other feature correlation and using better optimization methods \cite{taylor2013automatic,irie2016target,taylor2015multi}. However, these methods require an initial calibration provided elsewhere. Besides, the optimization is inefficient and can take several minutes.
Methods without an initial calibration have also been proposed in \cite{taylor2015motion,ishikawa2018lidar,huang2017extrinsic,chien2016visual,shi2019extrinsic,napier2013cross} using visual odometry. They align trajectories obtained from image and point cloud sequences. Nevertheless, their performance strongly depends on the result of trajectory estimation and they also have low efficiency.  

Meanwhile, as deep learning has witnessed huge progress in recent years, researchers have also been exploring methods using deep learning techniques.  \cite{schneider2017regnet,iyer2018calibnet} trained an end-to-end model that can correct the calibration within a rotation and translation boundary, while those methods are not universal and their performance depends heavily on the training dataset. There will exist lots of repetitive work such as data preparation once the type of LiDAR or camera changes.

Levinson \etal \cite{levinson2013automatic} proposed a novel calibration method for  online calibration. They suggested that discontinuities in laser scans correspond to edges in the image. The calibration result is given by optimizing a cost function that measures how well each laser point corresponds to an edge in the image. Unfortunately, this method is  vulnerable to noises and can only adjust the calibration in a small range. With the development of semantic segmentation techniques, Zhu \etal \cite{zhu2020online} argued that semantic features are more robust and effective than  edge features. They extracted cars from both the image and the point cloud, and the calibration is optimized through a similar cost function under semantic constraints. But their method still needs an initial calibration from other methods. Wang \etal \cite{wang2020soic} presented that an initial calibration can be computed by aligning SCs and solving the PnP problem. They also considered more semantic classes such as pedestrians and cyclists. However, the SCs often lie near the ground plane, which adds difficulty to solve the initial calibration. As a result, a robust initial calibration and an accurate final result would require tens of frames to have enough semantic constraints.
In this work, we propose to solve the initial calibration through the consistency of line features in a single frame of image and point cloud data.

\section{METHODOLOGY}
Given a single frame of image and point cloud data in road scenes, our method could automatically compute the extrinsic parameter between the corresponding camera and LiDAR. Firstly, a \textit{line feature extractor} could capture line pairs for both the image and the point cloud, which includes a lane detector to extract road lanes, and a pole detector to segment pole objects, such as street lights and telegraphy poles  on the side of the road. Then, several lines are fitted from the detected lanes and pole features, with which, a coarse calibration can be estimated by finding the correct line correspondences and solving the P3L problem. Finally, a cost function defined under the semantic constraints of detected lanes and poles is utilized to optimize the coarse calibration to the refined result. We introduce the details of all these processes in the following subsections.

\subsection{Problem Formulation}
The whole calibration process aligns a point cloud  and an image in road scenes, where the AV is driving on a straight and flat road with road lanes painted and poles on the side of the road. In our setting, the point cloud and the image are temporally synchronized, and we assume both the LiDAR and the camera are used for perceiving road scenes, therefore their field of view should have an overlap in the road area. We also assume that the LiDAR and the camera work under their desired lighting and weather conditions.

The point cloud is denoted as \textbf{$P^\mathrm{L}$} = \{$\mathbf{p}_1^\mathrm{L}, \mathbf{p}_2^\mathrm{L},..$\}, where $\mathrm{L}$ represents the LiDAR coordinate system. For each 3D point $\mathbf{p}_i^\mathrm{L} = (x_i,y_i,z_i)^T \in \mathbb{R}^3$, we denote its corresponding pixel coordinate on the image plane as $\mathbf{q}_i = (u_i,v_i)^T \in \mathbb{R}^2$. The point and the pixel can then be correlated by the calibration process in two steps. First, the point $\mathbf{p}_i^\mathrm{L}$ is transformed to  $\mathbf{p}_i^\mathrm{C} \in \mathbb{R}^3$, which is under the camera coordinate system $\mathrm{C}$. This is a rigid body transformation and can be represented as
\begin{equation}
\label{equ:extrinsic}
    \mathbf{p}_i^\mathrm{C} = \boldsymbol{\mathrm{R}}(\mathbf{r}) \cdot \mathbf{p}_i^\mathrm{L} + \boldsymbol{\mathrm{t}},
\end{equation}
where $\boldsymbol{\mathrm{R}}(\textbf{r})$ represents the rotation and can be parameterized by angle-axis representation $\mathbf{r}$, where $\frac{\mathbf{r}}{\lVert \mathbf{r}\Vert_2}$ is a unit vector representing the rotation axis and $\lVert \mathbf{r}\Vert_2$ is the rotation angle. $\boldsymbol{\mathrm{t}}$ represents the translation with $\boldsymbol{\mathrm{t}} = (t_x,t_y,t_z)^T$. The pair ($\mathbf{r},\mathbf{t}$) has six degrees of freedom~(6DoF) and is the extrinsic parameter between LiDAR and camera. Next, $\mathbf{p}_i^\mathrm{C}$ is projected onto the image plane through a projection function:
\begin{equation}
\label{equ:project}
    \boldsymbol{\mathcal{K}}: \mathbb{R}^3 \rightarrow \mathbb{R}^2, \mathbf{q}_i = \boldsymbol{\mathcal{K}} (\mathbf{p}_i^\mathrm{C}) \ .
\end{equation}
$\boldsymbol{\mathcal{K}}$ may vary with different projection models and can be defined by the camera intrinsic parameter such as focal length and lens distortion. 
In this work, we assume that the intrinsic parameter of camera is well calibrated and we focus on exploring the accurate extrinsic parameter $(\mathbf{r}, \boldsymbol{\mathrm{t}})$. 

\par 

\subsection{Line Feature Extractor}

\label{sec:lfe}

From the perspective of extracting line pairs with abundant semantic information, we propose the \textit{line feature extractor} to process both the image and the point cloud.
The static road lanes, lights, and telegraphy poles serve as the supplier of line features.
For the point cloud, we focus on utilizing the geometric priors and sensor characteristics to segment these targets.
For the image, the feature extraction is mainly based on semantic segmentation and contour refinement.

\subsubsection{Point Cloud Extraction}
For the point cloud, we will extract points from \textbf{$P^\mathrm{L}$} which belong to line feature.
First of all, $P^\mathrm{L}$ can be segmented into two subsets $P_{ground}^\mathrm{L}$ and $P_{object}^\mathrm{L}$ by performing a ground plane separation through RANSAC plane fitting\cite{fischler1981random}, where $P_{ground}^\mathrm{L}$ consists of all points that belong to the ground plane and $P_{object}^\mathrm{L}$ consists of points from elsewhere. The thickness of ground plane is set to 0.2m, which is the maximum elevation offset between any two points in $P_{ground}^{L}$.

Next, the points of the road lane $P_{lane}^\mathrm{L}$ are extracted from $P_{ground}^\mathrm{L}$ by utilizing an important characteristic of LiDAR sensor, intensity. Most road lanes are painted with high-reflectance materials to increase visibility in the dark, which gives them higher LiDAR signal response, \textit{i.e.}, higher intensity. Hence, we can extract the lane part by setting an intensity threshold $I_0$ and  points whose intensity values 
are larger than $I_0$ are kept.

Besides, in order to remove points from noise and other high-reflectance objects, we fit straight lines in the remaining points by RANSAC line fitting~\cite{fischler1981random}, which could reject outliers and merge line segments such as dotted lanes into one. Then, the distance $d_{min}$ from each point to its closest line could be easily calculated and the points whose $d_{min}$ is smaller than a threshold $D_0$ will be retained. Thus, the whole process could be summarized as follows,
\begin{equation}
\footnotesize	
    P_{lane}^\mathrm{L} = \{  \mathbf{p} \ | \ \mathbf{p}\in P_{ground}^\mathrm{L}, \ \mathbf{p} \veryshortarrow \mathrm{intensity} > I_0 \ \text{and} \ \mathbf{p} \veryshortarrow d_{min} < D_0 \}\ .
\end{equation}
We adjust $I_0$ to be $\mu_I + \sigma_I$ to adaptive extract $P_{lane}^\mathrm{L}$ in different  environments, where $\mu_I$ and $\sigma_I$ are the mean and standard deviation of intensity values for all points from $P_{ground}^\mathrm{L}$. And we set $D_0 = 0.3$ m which is twice the common width of road lane.

Afterwards, the points of poles $P_{pole}^\mathrm{L}$ can be extracted from $P_{object}^\mathrm{L}$. To improve robustness, we  rotate the $x-y$ plane of $\mathrm{L}$ to be parallel to the ground plane, and set the $x$ axis to be parallel to one line extracted in the ground plane. We call this new coordinate system the ground-parallel coordinate system $\mathrm{G}$.
These two coordinate systems, $\mathrm{G}$ and $\mathrm{L}$, have the same origin while the $Z$-axis of $\mathrm{G}$ is vertical to the ground plane computed in the previous step. Hence the $z$ coordinate of $\mathrm{G}$ corresponds to the ground elevation value of the points with respect to the elevation of the LiDAR. This transformation can be achieved through a rotation as
\begin{equation}
    \mathbf{p}_{object}^\mathrm{G} = \mathbf{R}_\mathrm{L}^\mathrm{G} \cdot \mathbf{p}_{object}^\mathrm{L} \ .
\end{equation}
We then construct a 2D grid $S$ onto the $x-y$ plane under $\mathrm{G}$. $S$ is constructed within the $x^\mathrm{G}$ boundary of $[0\text{ m}, 100\text{ m}]$ and the $y^\mathrm{G}$ boundary of $[-20\text{ m}, 20\text{ m}]$ in our experiment, to make sure only the side of the road is included. For each grid cell $s\in S$, it contains points in $P_{object}^\mathrm{G}$ that are inside the vertical pillar of $s$.
We denote $\boldsymbol{\mathcal{H}}_{\mathrm{max}}(s)$ to be the maximum elevation value in cell $s$, and only keep the grid cells whose $\boldsymbol{\mathcal{H}}_{\mathrm{max}}(s)$ are larger than an elevation threshold $H_1$. Thus, the set of points from all such grids $P_{S}^G$ can be presented as
\begin{equation}
    P_{S}^\mathrm{G} = \bigcup \Tilde{S}, \\\text{where}\ \Tilde{S}=\{  s\in S \ | \ \boldsymbol{\mathcal{H}}_{\mathrm{max}}(s) > H_1 \} \ .
\end{equation}

In addition, points with elevation smaller than a threshold $H_0$ are also discarded due to large noise near the ground plane, with which we can extract points of poles $P_{pole}^\mathrm{G}$:
\begin{equation}
    P_{pole}^\mathrm{L} = \{ ( \mathbf{R}_\mathrm{L}^\mathrm{G})^{-1}\cdot\mathbf{p}^\mathrm{G} \ |\  \mathbf{p}^\mathrm{G}\in\ P_{S}^\mathrm{G}\ \text{and}\ \mathbf{p}^\mathrm{G}  \veryshortarrow z > H_0 \ \} \ .
\end{equation}
In our experiment, we use $H_0 = -1 \ \text{m}$ and $H_1=3 \ \text{m}$. And grid cell length is set to $ 0.5 \ \text{m}$, ensuring that at most one pole appears in a single cell. 
 

\subsubsection{Image Segmentation}
In the field of the image, BiSeNet-V2\cite{yu2018bisenet} has been widely used as one of the state-of-the-art semantic segmentation networks with superior performance and real-time inference speed. We keep the architecture of BiSeNet-V2 to get preliminary segmentation results. However, poles and lanes are very thin objects and it is hard to capture their contours accurately using a network. Therefore, we further refine the contours using a Dense CRF operator\cite{krahenbuhl2011efficient}, which matches pixels with similar colors as well as close positions.
As shown in Fig.~\ref{fig:crf}, the contours of poles are more compact and realistic after the CRF operator.  


\figcrf

Consequently, the pixels from poles $Q_{pole}$ and from road lanes $Q_{lane}$ can be directly obtained from the class label ``pole" and ``road lane". Combining the segmentation results, we can obtain two binary masks $\boldsymbol{\mathcal{M}}_{line}: \mathbb{R}^2 \rightarrow \{0,1\}, line\in\{pole,lane\}$  on the pixel coordinate defined as follows: 
\begin{equation} 
\begin{split}
    \boldsymbol{\mathcal{M}}_{line}(\mathbf{q}) :=
    \begin{cases}
        1& \mathbf{q}\in Q_{line}\\
        0& \mathrm{otherwise}
    \end{cases}.
\end{split}
\end{equation}

\subsubsection{Cost Function}
After line features are extracted from both the image and the point cloud, we propose a cost function that measures how well the image and the point cloud is correlated given an extrinsic parameter $(\mathbf{r},\mathbf{t})$ using semantic constraints.  

First of all, similar to \cite{zhu2020online}, we apply an inverse distance transformation (IDT) to the mask $ \boldsymbol{\mathcal{M}}_{line}$ to avoid duplicate local maxima during later optimization. The resulting height map  $\boldsymbol{\mathcal{H}}_{line}, line\in\{pole,lane\}$ is defined as follows:
\begin{equation}
      \boldsymbol{\mathcal{H}}_{line}(\mathbf{q}) :=
    \begin{cases}
\max\limits_{\mathbf{s} \in \mathbb{R}^2 \setminus Q_{line}} \gamma_0^{\lVert \mathbf{q}-\mathbf{s} \Vert_1}& \mathbf{q}\in Q_{line}\\
\max\limits_{\mathbf{s} \in Q_{line}}  \gamma_1^{\lVert \mathbf{q}-\mathbf{s} \Vert_1} & \mathbf{q}\in \mathbb{R}^2 \setminus Q_{line}
\end{cases}  .
\end{equation}

Next, we propose our cost function $ \boldsymbol{\mathcal{J}}: (\mathbf{r},\mathbf{t}) \rightarrow \mathbb{R}$, which represents the consistency between the projecting pixels of $P_{lane}$ and $P_{pole}$ and their corresponding masks in the image. The cost function $ \boldsymbol{\mathcal{J}}$ is defined as
\begin{equation}
\begin{split}
    \boldsymbol{\mathcal{J}}   =  \sum\limits_{line \in \{pole, \ lane \} }  & \displaystyle\frac{\sum\limits_{\mathbf{p} \in P_{line}^\mathrm{L} } \boldsymbol{\mathcal{H}}_{line}\circ \boldsymbol{\mathcal{K}}(\mathbf{R}(\mathbf{r}) \mathbf{p} +\mathbf{t})}{|P_{line}^\mathrm{L} |}  
\end{split},
\end{equation}
which uses (\ref{equ:extrinsic}) and (\ref{equ:project}) mentioned above for projecting points. $|P_{line}^\mathrm{L} |$  refers to the number of points in $P_{line}^\mathrm{L}$  and is used to balance the cost between poles and lanes. The larger the cost function is, the better the semantic features from two data domains match.  

Compared to the cost function proposed in \cite{wang2020soic}, our cost function is cheaper to compute as points from line features are fewer than other semantic classes like cars. Moreover, the line features are from different spatial positions rather than clustered near the ground plane, which can provide stronger spatial constraints and increase the robustness of calibration.

\subsection{Coarse Calibration}
\label{sec:coarse}
As stated before-head, most target-less calibration methods assume an initial calibration is available. On the contrary, our method could provide a coarse calibration from line correspondences which acts as an initialization to the later optimization process as shown in Fig.~\ref{fig:p3l}.

\figpnl

We first fit lines of road lanes and poles in the image as ($l_{lane}^\mathrm{I}$, $l_{pole}^\mathrm{I}$)  and  in the point cloud as ($l_{lane}^\mathrm{L}$, $l_{pole}^\mathrm{L}$) respectively. $l_{lane}^\mathrm{L}$ and $l_{pole}^\mathrm{L}$ can be obtained by RANSAC line fitting in $P_{lane}^\mathrm{L}$ and $P_{pole}^\mathrm{L}$. For $l_{pole}^\mathrm{I}$ and  $l_{lane}^\mathrm{I}$, we extract them through Hough line transform \cite{duda1972use} on $\boldsymbol{\mathcal{M}}_{lane}$ and $\boldsymbol{\mathcal{M}}_{pole}$. 

After the line equations are computed, we establish in total three-line correspondences including two-lane correspondences and one-pole correspondence in the image and the point cloud to solve the P3L problem and  obtain the coarse calibration. To find the correct line correspondences,  we first select $l_{lane1}^\mathrm{I}$, $l_{lane2}^\mathrm{I}$, and $l_{pole1}^\mathrm{I}$ with the largest pixel area from the image and keep them fixed. Then we enumerate all  $l_{lane}^\mathrm{L}$ and $l_{pole}^\mathrm{L}$ in the point cloud and solve the P3L problem, which yields a coarse calibration for each possible correspondence. Suppose we have $n_1$ $l_{lane}^\mathrm{L}$ and $n_2$ $l_{pole}^\mathrm{L}$, then we will obtain $n_1(n_1-1)n_2$ candidates for the coarse calibration.  Finally we evaluate each course calibration candidate using the cost function $\boldsymbol{\mathcal{J}}$, and  one with the largest cost should be associated with the correct correspondence.

In a general setting, the P3L problem involves solving an eighth-order equation. Since lanes are often parallel to each other, the problem can be greatly simplified by assuming  $l_{lane1}^\mathrm{L} \parallel l_{lane2}^\mathrm{L}$. To exploit this relationship, we first transform $l_{lane}^\mathrm{L}$ and $l_{pole}^\mathrm{L}$ to the ground-parallel coordinate system $\mathrm{G}$ as an intermediate step, where the following relationship holds: 

\begin{equation}
    \mathbf{p}^\mathrm{C} = \mathbf{R}_\mathrm{G}^\mathrm{C}  \cdot \mathbf{R}_\mathrm{L} ^\mathrm{G}  \cdot \mathbf{p}^\mathrm{L}  + \mathbf{t}_\mathrm{G} ^\mathrm{C} \ .
\end{equation}

  As $\mathbf{R}_\mathrm{L} ^\mathrm{G}$ is already known, we compute $\mathbf{R}_\mathrm{G} ^\mathrm{C}$ and $\mathbf{t}_\mathrm{G} ^\mathrm{C}$ using the method introduced in  \cite{xu2016pose}.   $\mathbf{R}_\mathrm{G} ^\mathrm{C}$ can be parameterized as three consecutive rotations:
 \begin{equation}
     \mathbf{R}_\mathrm{G} ^ \mathrm{C} = \mathbf{R}'\cdot  \mathrm{Rot}(X,\alpha)\cdot \mathrm{Rot}(Z,\beta) \ ,
 \end{equation}
 where $\mathbf{R}'$ is an arbitrary rotation matrix whose first column is equal to the normal $\mathbf{n}^\mathrm{C}$ of the plane back-projected by $l_{pole}^{\mathrm{I}}$, and $\mathrm{Rot}(X,\alpha)$ and $\mathrm{Rot}(Z,\beta)$ represent a rotation around the current $X$-axis and the current $Z$-axis respectively. In this setting, the order of the equation for $\alpha$ and $\beta$ is reduced to 2, and $\mathbf{R}_\mathrm{G} ^\mathrm{C}$ can be solved subsequently. Next  $\mathbf{t}_G^C$ can be obtained by solving the following system of linear equations:
 \begin{equation}
     {\mathbf{n}_i^\mathrm{C}}^T \cdot  \mathbf{t}_\mathrm{G}^\mathrm{C} =  -{\mathbf{n}_i^\mathrm{C}}^T \cdot \mathbf{R}_\mathrm{G}^\mathrm{C}\cdot  \mathbf{R}_\mathrm{L}^\mathrm{G} \cdot \mathbf{p}_i^\mathrm{L} \indent (i = 1,2,3) \ ,
 \end{equation}
where $\mathbf{p}_i^L$ is the coordinate of any point on the $i$-th line $l_i^\mathrm{L}$ and $\mathbf{n}_i^\mathrm{C}$ is the normal of the plane back-projected by $l_i^\mathrm{I}$. Finally, the coarse calibration  $(\mathbf{r_0},\mathbf{t_0})$ is obtained, with 
\begin{equation}
    \mathbf{R}(\mathbf{r_0}) = \mathbf{R}_\mathrm{G}^\mathrm{C}\cdot  \mathbf{R}_\mathrm{L}^\mathrm{G},  \mathbf{t_0} = \mathbf{t}_\mathrm{G}^\mathrm{C} \ .
\end{equation}

\subsection{Calibration Refinement}
As the coarse calibration only uses three-line correspondences and lanes are not perfectly parallel in real world, we consider all line features extracted in Sec. \ref{sec:lfe} and discard the parallel assumption to further refine the coarse calibration by directly optimizing the cost function $\boldsymbol{\mathcal{J}}$, and the  refined calibration  $(\mathbf{\hat{r}},\mathbf{\hat{t}})$ should maximize the cost function:
\begin{equation}
    (\mathbf{\hat{r}},\mathbf{\hat{t}}) = \mathrm{arg}\max\limits_{(\mathbf{r},\mathbf{t})}  \boldsymbol{\mathcal{J}}(\mathbf{R}(\mathbf{r}),\mathbf{t}) \ .
\end{equation}
$\boldsymbol{\mathcal{J}}$ is non-convex and can be hardly optimized by  convex optimization techniques. However, as our coarse calibration can already provide a reasonable and robust initial estimation as shown in Sec. \ref{sec:qr}, we first initialize the refinement with the coarse calibration $(\mathbf{r_0}, \mathbf{t_0})$, and subsequently $\boldsymbol{\mathcal{J}}$ is optimized by random searching the parameter space around the current optimal parameter. For translation, we sample $\delta \mathbf{t}$ as three independent components $(\delta t_x,\delta t_y,\delta t_z)$ between $[-t_0, t_0]$.  For rotation, we sample $\delta \mathbf{r}$ following axis-angle representation where the axis is sampled on the unit sphere and the angle is sampled within a boundary $[-\theta_0, \theta_0]$.  The calibration refinement algorithm is shown in Algorithm~\ref{alg:1}. In our implementation, we set $t_0$ to be $1$ m and $\theta_0$ to be $0.1$\degree. The initial step size $\eta$ and the final step size $\eta_0$ are set to be $1$ and $0.001$ respectively. The decay factor $k$ is $0.1$ and the search count is limited to $10000$. 

\alg 

\section{EXPERIMENTS}
\subsection{Experiment Settings} 
To evaluate the performance of CRLF, experiments are conducted on two sets of data. The first one is the KITTI dataset \cite{Geiger2013IJRR}, which is commonly used as a benchmark for studies related to autonomous driving. We adopt the KITTI raw dataset containing original sensor data and calibration files. The point cloud data are obtained from a Velodyne HDL-64 LiDAR, and images are from the rectified left RGB camera with a resolution of $1242\times375$. The camera intrinsic parameter from the calibration file is taken as ground truth and the extrinsic parameter serves as a reference for evaluating the accuracy of CRLF. Our data is selected from various road scenes at intervals of 10 frames to ensure that there is enough disparity between any two frames. In total, 100 frames are collected for evaluation from 3 driving sequences (09/26/2011 drive 0015, 0036, and 0101).

Additionally, the test data collected by our self-driving cars are used to better demonstrate the generality of CRLF. The images are from a front center camera with a resolution of $1920\times1200$ and the point cloud data is collected from a 64-beam Hesai LiDAR.  We also select 100 frames of data to compare with the reference extrinsic parameter, which is obtained through manually selecting feature point correspondences between the point cloud and the image to solve the PnP problem and is carefully tuned under several different scenes.

To evaluate the performance of CRLF with respect to the reference calibration, we separately measure the error for translation and rotation. Translation error is calculated as the euclidean distance between the two translations as
\begin{equation}
    \Delta t = \lVert \hat{\mathbf{t}} - \mathbf{t }\Vert_2.
\end{equation}
Rotation error is
the geodesic distance
based on angle-axis representation, which is defined as:
\begin{equation}
   \Delta \theta =   \lVert \log( \mathbf{R}(\mathbf{\hat{e}}) \mathbf{R}(\mathbf{e})^T) \Vert_2 \ ,
\end{equation}
where the $\lVert. \Vert$ gives the magnitude of the angle between any two rotations. We also calculate the mean absolute error (MAE) for the three components of translation, namely $\Delta t_x$, $\Delta t_y$, and $\Delta t_z$, as well as the MAE for the three Euler angles $\Delta$\textit{roll}, $\Delta$\textit{pitch}, and $\Delta$\textit{yaw}, which follow the $ZYX$ representation.

Our method is implemented in C++ on a desktop computer with an Intel Core i7-8700 CPU and a Nvidia 1660 GPU. The average computation time for each step of the calibration procedure  is recorded in Table \ref{table:efficiency}. The whole procedure only takes around $0.3\text{ s}$ to run. Moreover, as our method requires a single pair of point cloud and image and is fully-automatic, we conclude that our method is highly efficient and user-friendly to be deployed on a broad scale of AVs.
\figqualitativee
 
\tableEfficiency
\tableResult
\subsection{Quantitative Results}
\label{sec:qr}
We first run coarse calibration and compute the error of CRLF on both datasets. The MAE is shown in Table \ref{table:result} and the error distribution is visualized in Fig. \ref{fig:coarse}. On both datasets, the coarse calibration can give a reasonable initial value with the maximum translation error less than $0.5$ m and the maximum rotation error less than $3\degree$. We can see that about $80\%$ of the translation errors is less than $0.3$ m on KITTI and $0.2$ m on our own dataset, and $80\%$ of the rotation errors is less than $2.3\degree$ and $2\degree$ respectively.


\figcoarse
 
After the coarse calibration, the calibration refinement is applied and the final result is compared to the reference calibration. The MAE is also listed in Table \ref{table:result} and Fig. \ref{fig:refine} shows the cumulative error distribution of the refined calibration, which has a significant improvement compared to the coarse calibration on both datasets.

To further demonstrate the robustness of our refinement procedure, we challenge the procedure through applying random transformations to the reference calibration value in the range of $[-1\text{ m}, 1\text{ m}]$  for translation and $[-6\degree, 6\degree]$ for rotation, which are twice the maximum absolute error of coarse calibration respectively. This transformation is applied to each data, and the MAE of the refined calibration versus the miscalibration error of each initial transformation is plotted in Fig. \ref{fig:converge}. The MAE only slightly increases with the miscalibration error, and this may be caused by the optimization getting stuck at a local maximum occasionally. $\Delta$\textit{yaw} is around $0.5$\degree on all the miscalibration error, we infer that the constraint information for the global $y$-axis on image plane is less than others due to the direction of poles and lanes.
Nevertheless, the overall convergence performance is stable, and the initial error can be reduced to up to one-tenth after the refinement.

\figrefine

\figconverge

\subsection{Qualitative Results}

To better visualize the performance of CRLF, the point cloud is projected to the image plane using the extrinsic parameter. Results on our in-house dataset and KITTI dataset are shown in Fig.~\ref{fig:qualitative}. The coarse calibration can already align line features roughly to their corresponding masks in Fig.~\ref{fig:qualitative}(b). However, since three-line correspondences only provide a spatial constraint, there is still misalignment at global semantic level as shown in the pink box. Fig.~\ref{fig:qualitative}(c) shows that the refinement procedure corrects the misalignment and projects more points within the mask compared to Fig.~\ref{fig:qualitative}(b).
After the refinement, all line features align well, and Fig.~\ref{fig:qualitative}(d) demonstrates the overall quality of CRLF, where objects like cars and trees are also well-aligned.

\section{CONCLUSIONS}
In this paper, we propose a novel fully-automatic method, CRLF, for extrinsic parameter calibration between a LiDAR and a camera.  We leverage static straight-line-shaped objects including lanes and poles to enrich our potential targets for calibration. A line feature extractor is presented to extract straight-line features in road scenes for both the image and the point cloud. The line features provide not only enough spatial constraints to robustly estimate an accurate initial calibration but also abundant semantic information for further refinement.
Qualitative and quantitative results demonstrate the robustness and effectiveness of CRLF. The experiments also show CRLF's promising potential to be deployed on a large scale of AVs for companies and users in the real world.



As CRLF can be applied to various scenarios, its performance depends heavily on the line feature extractor, which may suffer from poor environmental conditions, e.g., fog, rain, and night, that affect sensor measurements, as well as unexpected objects that are wrongly extracted by the segmentation models. Moreover, as line-shaped objects can already provide enough spatial constraints, we look forward to using more geometric-shaped objects in different ranges and positions to further improve the performance.

\bibliographystyle{IEEEtran}
\bibliography{egbib}

\begin{thebibliography}{10}
\providecommand{\url}[1]{#1}
\csname url@rmstyle\endcsname
\providecommand{\newblock}{\relax}
\providecommand{\bibinfo}[2]{#2}
\providecommand\BIBentrySTDinterwordspacing{\spaceskip=0pt\relax}
\providecommand\BIBentryALTinterwordstretchfactor{4}
\providecommand\BIBentryALTinterwordspacing{\spaceskip=\fontdimen2\font plus
\BIBentryALTinterwordstretchfactor\fontdimen3\font minus
  \fontdimen4\font\relax}
\providecommand\BIBforeignlanguage[2]{{%
\expandafter\ifx\csname l@#1\endcsname\relax
\typeout{** WARNING: IEEEtran.bst: No hyphenation pattern has been}%
\typeout{** loaded for the language `#1'. Using the pattern for}%
\typeout{** the default language instead.}%
\else
\language=\csname l@#1\endcsname
\fi
#2}}

\bibitem{geiger2012automatic}
A.~Geiger, F.~Moosmann, {\"O}.~Car, and B.~Schuster, ``Automatic camera and
  range sensor calibration using a single shot,'' in \emph{2012 IEEE
  International Conference on Robotics and Automation}, 2012, pp. 3936--3943.

\bibitem{wang2017reflectance}
W.~Wang, K.~Sakurada, and N.~Kawaguchi, ``Reflectance intensity assisted
  automatic and accurate extrinsic calibration of 3d lidar and panoramic camera
  using a printed chessboard,'' \emph{Remote Sensing}, vol.~9, no.~8, p. 851,
  2017.

\bibitem{zhou2018automatic}
L.~Zhou, Z.~Li, and M.~Kaess, ``Automatic extrinsic calibration of a camera and
  a 3d lidar using line and plane correspondences,'' in \emph{2018 IEEE/RSJ
  International Conference on Intelligent Robots and Systems (IROS)}, 2018, pp.
  5562--5569.

\bibitem{an2020geometric}
P.~An, T.~Ma, K.~Yu, B.~Fang, J.~Zhang, W.~Fu, and J.~Ma, ``Geometric
  calibration for lidar-camera system fusing 3d-2d and 3d-3d point
  correspondences,'' \emph{Optics Express}, vol.~28, no.~2, pp. 2122--2141,
  2020.

\bibitem{pusztai2017accurate}
Z.~Pusztai and L.~Hajder, ``Accurate calibration of lidar-camera systems using
  ordinary boxes,'' in \emph{Proceedings of the IEEE International Conference
  on Computer Vision Workshops}, 2017, pp. 394--402.

\bibitem{levinson2013automatic}
J.~Levinson and S.~Thrun, ``Automatic online calibration of cameras and
  lasers.'' in \emph{Robotics: Science and Systems}, vol.~2, 2013, p.~7.

\bibitem{zhu2020online}
Y.~Zhu, C.~Li, and Y.~Zhang, ``Online camera-lidar calibration with sensor
  semantic information,'' in \emph{2020 IEEE International Conference on
  Robotics and Automation (ICRA)}, 2020, pp. 4970--4976.

\bibitem{wang2020soic}
W.~Wang, S.~Nobuhara, R.~Nakamura, and K.~Sakurada, ``Soic: Semantic online
  initialization and calibration for lidar and camera,'' \emph{arXiv preprint
  arXiv:2003.04260}, 2020.

\bibitem{park2014calibration}
Y.~Park, S.~Yun, C.~S. Won, K.~Cho, K.~Um, and S.~Sim, ``Calibration between
  color camera and 3d lidar instruments with a polygonal planar board,''
  \emph{Sensors}, vol.~14, no.~3, pp. 5333--5353, 2014.

\bibitem{velas2014calibration}
M.~Vel'as, M.~{\v{S}}pan{\v{e}}l, Z.~Materna, and A.~Herout, ``Calibration of
  rgb camera with velodyne lidar,'' in \emph{WSCG 2014 Communication Papers
  Proceedings}, 2014, pp. 135--144.

\bibitem{cai2020novel}
H.~Cai, W.~Pang, X.~Chen, Y.~Wang, and H.~Liang, ``A novel calibration board
  and experiments for 3d lidar and camera calibration,'' \emph{Sensors},
  vol.~20, no.~4, p. 1130, 2020.

\bibitem{pandey2012automatic}
G.~Pandey, J.~R. McBride, S.~Savarese, and R.~M. Eustice, ``Automatic
  targetless extrinsic calibration of a 3d lidar and camera by maximizing
  mutual information.'' in \emph{AAAI}, 2012.

\bibitem{taylor2013automatic}
Z.~Taylor and J.~Nieto, ``Automatic calibration of lidar and camera images
  using normalized mutual information,'' in \emph{2013 IEEE International
  Conference on Robotics and Automation (ICRA)}, 2013.

\bibitem{irie2016target}
K.~Irie, M.~Sugiyama, and M.~Tomono, ``Target-less camera-lidar extrinsic
  calibration using a bagged dependence estimator,'' in \emph{2016 IEEE
  International Conference on Automation Science and Engineering (CASE)}, 2016,
  pp. 1340--1347.

\bibitem{taylor2015multi}
Z.~Taylor, J.~Nieto, and D.~Johnson, ``Multi-modal sensor calibration using a
  gradient orientation measure,'' \emph{Journal of Field Robotics}, vol.~32,
  no.~5, pp. 675--695, 2015.

\bibitem{taylor2015motion}
Z.~Taylor and J.~Nieto, ``Motion-based calibration of multimodal sensor
  arrays,'' in \emph{2015 IEEE International Conference on Robotics and
  Automation (ICRA)}, 2015, pp. 4843--4850.

\bibitem{ishikawa2018lidar}
R.~Ishikawa, T.~Oishi, and K.~Ikeuchi, ``Lidar and camera calibration using
  motions estimated by sensor fusion odometry,'' in \emph{2018 IEEE/RSJ
  International Conference on Intelligent Robots and Systems (IROS)}, 2018, pp.
  7342--7349.

\bibitem{huang2017extrinsic}
K.~Huang and C.~Stachniss, ``Extrinsic multi-sensor calibration for mobile
  robots using the gauss-helmert model,'' in \emph{2017 IEEE/RSJ International
  Conference on Intelligent Robots and Systems (IROS)}, 2017, pp. 1490--1496.

\bibitem{chien2016visual}
H.-J. Chien, R.~Klette, N.~Schneider, and U.~Franke, ``Visual odometry driven
  online calibration for monocular lidar-camera systems,'' in \emph{2016 23rd
  International conference on pattern recognition (ICPR)}, 2016, pp.
  2848--2853.

\bibitem{shi2019extrinsic}
C.~Shi, K.~Huang, Q.~Yu, J.~Xiao, H.~Lu, and C.~Xie, ``Extrinsic calibration
  and odometry for camera-lidar systems,'' \emph{IEEE Access}, vol.~7, pp.
  120\,106--120\,116, 2019.

\bibitem{napier2013cross}
A.~Napier, P.~Corke, and P.~Newman, ``Cross-calibration of push-broom 2d lidars
  and cameras in natural scenes,'' in \emph{2013 IEEE International Conference
  on Robotics and Automation}, 2013, pp. 3679--3684.

\bibitem{schneider2017regnet}
N.~Schneider, F.~Piewak, C.~Stiller, and U.~Franke, ``Regnet: Multimodal sensor
  registration using deep neural networks,'' in \emph{2017 IEEE intelligent
  vehicles symposium (IV)}, 2017, pp. 1803--1810.

\bibitem{iyer2018calibnet}
G.~Iyer, R.~K. Ram, J.~K. Murthy, and K.~M. Krishna, ``Calibnet: Geometrically
  supervised extrinsic calibration using 3d spatial transformer networks,'' in
  \emph{2018 IEEE/RSJ International Conference on Intelligent Robots and
  Systems (IROS)}, 2018, pp. 1110--1117.

\bibitem{fischler1981random}
M.~A. Fischler and R.~C. Bolles, ``Random sample consensus: a paradigm for
  model fitting with applications to image analysis and automated
  cartography,'' \emph{Communications of the ACM}, vol.~24, no.~6, pp.
  381--395, 1981.

\bibitem{yu2018bisenet}
C.~Yu, J.~Wang, C.~Peng, C.~Gao, G.~Yu, and N.~Sang, ``Bisenet: Bilateral
  segmentation network for real-time semantic segmentation,'' in \emph{2018
  European conference on computer vision (ECCV)}, 2018, pp. 325--341.

\bibitem{krahenbuhl2011efficient}
P.~Kr{\"a}henb{\"u}hl and V.~Koltun, ``Efficient inference in fully connected
  crfs with gaussian edge potentials,'' in \emph{Advances in neural information
  processing systems}, 2011, pp. 109--117.

\bibitem{duda1972use}
R.~O. Duda and P.~E. Hart, ``Use of the hough transformation to detect lines
  and curves in pictures,'' \emph{Communications of the ACM}, vol.~15, no.~1,
  pp. 11--15, 1972.

\bibitem{xu2016pose}
C.~Xu, L.~Zhang, L.~Cheng, and R.~Koch, ``Pose estimation from line
  correspondences: A complete analysis and a series of solutions,'' \emph{IEEE
  transactions on pattern analysis and machine intelligence}, vol.~39, no.~6,
  pp. 1209--1222, 2016.

\bibitem{Geiger2013IJRR}
A.~Geiger, P.~Lenz, C.~Stiller, and R.~Urtasun, ``Vision meets robotics: The
  kitti dataset,'' \emph{International Journal of Robotics Research (IJRR)},
  vol.~32, no.~11, p. 1231–1237, 2013.

\end{thebibliography}

\end{document}